\documentclass[final]{cvpr}

\usepackage{times}
\usepackage{epsfig}
\usepackage{graphicx}
\usepackage{amsmath}
\usepackage{amssymb}

\usepackage[square,numbers]{natbib}
\usepackage[utf8]{inputenc} 
\usepackage[T1]{fontenc}    
\usepackage{url}            
\usepackage{booktabs}       
\usepackage{amsfonts}       
\usepackage{nicefrac}       
\usepackage{microtype}      
\usepackage{numprint}       
\usepackage{lipsum}
\usepackage{amsmath}
\usepackage{dsfont}
\usepackage{float}
\usepackage{graphicx}
\usepackage{caption}
\usepackage{subcaption}
\usepackage{subfiles}
\usepackage{placeins}
\usepackage{mathtools}
\usepackage{cuted}
\usepackage{listings,multicol}
\usepackage{amsmath}
\usepackage[ruled,vlined]{algorithm2e}

\usepackage{tabularx}
\usepackage{array}
    \newcolumntype{L}{>{\raggedright\arraybackslash}X}
\usepackage{multirow}
\usepackage{makecell}
\usepackage{color}

\usepackage[utf8]{inputenc}
\usepackage[english]{babel}

\newtheorem{theorem}{Theorem}[section]

\newtheorem{lemma}[theorem]{Lemma}

\definecolor{mygreen}{rgb}{0,0.6,0}
\definecolor{mygray}{rgb}{0.5,0.5,0.5}
\definecolor{mymauve}{rgb}{0.58,0,0.82}
\allowdisplaybreaks

\usepackage{listings}
\usepackage{xcolor}

\definecolor{codegreen}{rgb}{0,0.6,0}
\definecolor{codegray}{rgb}{0.5,0.5,0.5}
\definecolor{codepurple}{rgb}{0.58,0,0.82}
\definecolor{backcolour}{rgb}{0.95,0.95,0.92}

\lstdefinestyle{mystyle}{
    backgroundcolor=\color{backcolour},   
    commentstyle=\color{codegreen},
    keywordstyle=\color{magenta},
    numberstyle=\tiny\color{codegray},
    stringstyle=\color{codepurple},
    basicstyle=\ttfamily\footnotesize,
    breakatwhitespace=false,         
    breaklines=true,                 
    captionpos=b,                    
    keepspaces=true,                 
    numbers=none,                    
    numbersep=5pt,                  
    showspaces=false,                
    showstringspaces=false,
    showtabs=false,                  
    tabsize=2
}

\DeclareMathOperator*{\argmax}{arg\,max}

\DeclareMathOperator*{\softmax}{softmax}
\DeclareMathOperator*{\sigmoid}{sigmoid}
\newcommand{\brac}[1]{\left[#1\right]}
\newcommand{\vtab}{VTAB\textsubscript{1K}}

\newcommand{\parencurly}[1]{\left\{#1\right\}}

\usepackage[pagebackref=true,breaklinks=true,colorlinks,bookmarks=false]{hyperref}

\def\cvprPaperID{8309} 
\def\confYear{CVPR 2021}

\begin{document}

\title{Correlated Input-Dependent Label Noise in Large-Scale Image Classification}

\author{Mark Collier\\
Google AI\\
{\tt\small markcollier@google.com}
\and
Basil Mustafa\\
Google AI\\
{\tt\small basilm@google.com}
\and
Efi Kokiopoulou\\
Google AI\\
{\tt\small kokiopou@google.com}
\and
Rodolphe Jenatton\\
Google AI\\
{\tt\small rjenatton@google.com}
\and
Jesse Berent\\
Google AI\\
{\tt\small jberent@google.com}
}

\maketitle

\begin{abstract}
   Large scale image classification datasets often contain noisy labels. We take a principled probabilistic approach to modelling input-dependent, also known as heteroscedastic, label noise in these datasets. We place a multivariate Normal distributed latent variable on the final hidden layer of a neural network classifier. The covariance matrix of this latent variable, models the aleatoric uncertainty due to label noise. We demonstrate that the learned covariance structure captures known sources of label noise between semantically similar and co-occurring classes. Compared to standard neural network training and other baselines, we show significantly improved accuracy on Imagenet ILSVRC 2012 79.3\% (+ 2.6\%), Imagenet-$21k$ 47.0\% (+ 1.1\%) and JFT 64.7\% (+ 1.6\%). We set a new state-of-the-art result on WebVision 1.0 with 76.6\% top-1 accuracy. These datasets range from over 1M to over 300M training examples and from $1k$ classes to more than $21k$ classes. Our method is simple to use, and we provide an implementation that is a drop-in replacement for the final fully-connected layer in a deep classifier.
\end{abstract}

\FloatBarrier

\section{Introduction}

Image classification datasets with many classes and large training sets often have noisy labels~\cite{beyer2020we,li2017webvision}. For example, Imagenet contains many visually similar classes that are hard for human annotators to distinguish~\cite{deng2009imagenet,beyer2020we}. Datasets such as WebVision where labels are generated automatically by looking at co-occuring text to images on the Web, contain label noise as this automated process is not 100\% reliable~\cite{li2017webvision}.

\begin{figure}
     \centering
     \begin{subfigure}[b]{0.18\textwidth}
         \centering
         \includegraphics[width=\textwidth]{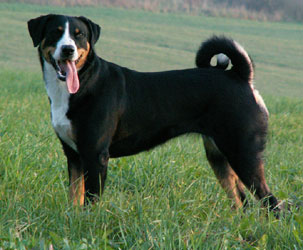}
     \end{subfigure}
     \begin{subfigure}[b]{0.168\textwidth}
         \centering
         \includegraphics[width=\textwidth]{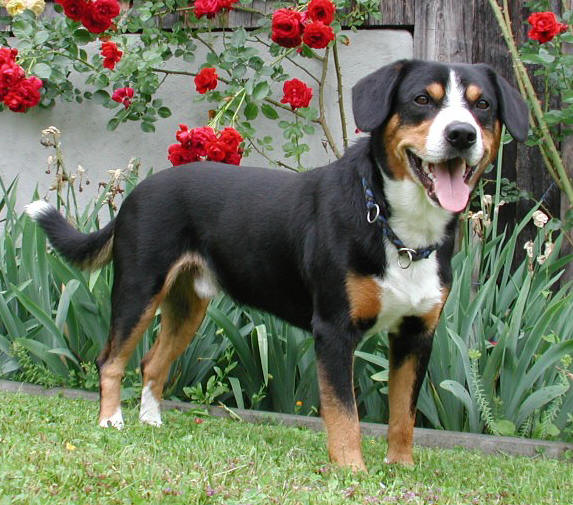}
     \end{subfigure}
        \caption{Spot the difference? An Appenzeller (left) and EntleBucher (right). Two visually similar Imagenet classes our method \textit{learns} have highly correlated label noise (average validation set covariance of -0.24) given only the standard Imagenet ILSVRC12 training labels.}
        \label{fig:doggos}
\end{figure}

A wide range of techniques for classification under label noise already exist~\cite{lee2018cleannet,MentorNet.2018,guo2018curriculumnet,saxena2019data,jiang2020beyond,cao2020heteroskedastic,collier2020analysis,reed6596training,CoTeaching.2018}. 
When an image is mis-labeled it is more likely that it gets confused with other related classes, rather than a random class \cite{beyer2020we}. Therefore it is important to take inter-class correlation into account when modelling label noise in image classification.

We take a principled probabilistic approach to modelling label noise. We assume a generative process for noisy labels with a multivariate Normal distributed latent variable at the final hidden layer of a neural network classifier. The mean and covariance parameters of this Normal distribution are input-dependent (aka heteroscedastic), being computed from a shared representation of the input image. By modelling the inter-class noise correlations our method can learn which class pairs are substitutes or commonly co-occur, resulting in noisy labels. See Fig.\ (\ref{fig:doggos}) for an example of two Imagenet classes which our model learns have correlated label noise.

We evaluate our method on four large-scale image classification datasets, Imagenet ILSVRC12 and Imagenet-$21k$~\cite{deng2009imagenet}, WebVision 1.0~\cite{li2017webvision} and JFT~\cite{distillation2015}. These datasets range from over 1M training examples (ILSVRC12) to 300M training examples (JFT) and from $1k$ classes (ILSVRC12 \& WebVision) to over $21k$ classes (Imagenet-$21k$). We demonstrate improved accuracy and negative log-likelihood on all datasets relative to (a) standard neural network training, (b) methods which only model the diagonal of the covariance matrix and (c) methods from the noisy labels literature.

We evaluate the effect of our probabilistic label noise model on the representations learned by the network. We show that our method, when pre-trained on JFT, learns image representations which transfer better to the 19 datasets from the Visual Task Adaptation Benchmark (VTAB) \cite{zhai2020largescale}.

\paragraph{Contributions.} In summary our contributions are:
\begin{enumerate}
    \itemsep0em 
    \item A new method which models inter-class correlated label noise and scales to large-scale datasets.
    \item We evaluate our method on four large-scale image classification datasets, showing significantly improved performance compared to standard neural network training and diagonal covariance methods.
    \item We demonstrate that the learned covariance matrices model correlations between semantically similar or commonly co-occurring classes.
    \item On VTAB our method learns more general representations which transfer better to 19 downstream tasks.
\end{enumerate}

\section{Method}
\label{sec:method}

In many datasets label noise is not uniform across the input space, some types of examples have more noise than others. We build upon prior work on probabilistic modelling of noisy labels~\cite{kendall2017uncertainties,collier2020analysis} by assuming a heteroscedastic latent variable generative process for our labels. This generative process leads to two main challenges while computing its resulting likelihood: (a) the intractable marginalization over the latent variables which we estimate via Monte Carlo integration and (b) an $\argmax$ in the generative process which we approximate with a temperature parameterized $\softmax$.

\paragraph{Generative process.}

Suppose there is some latent vector of utility $\mathbf{u}(\mathbf{x}) \in \mathbf{R}^K$, where  $K$ is the number of classes associated with each input $\mathbf{x}$.
This utility is the sum of a deterministic reference utility $\boldsymbol{\mu}(\mathbf{x})$ and an unobserved stochastic component $\boldsymbol{\epsilon}$. A label is generated by sampling from the utility and taking the $\argmax$, i.e. class $c$ is the label if its associated utility is greater than the utility for all other classes $\iff y = \argmax_{j \in [K]} \mathbf{u}_j(\mathbf{x})$:

\begin{equation}
\begin{split}
    \mathbf{u}(\mathbf{x}) &= \boldsymbol{\mu}(\mathbf{x}) + \boldsymbol{\epsilon} \\
    p_c &= P(y = c | \mathbf{x}) = P(\argmax_{j \in [K]} \mathbf{u}_j(\mathbf{x}) = c) \\
    &= \int \mathds{1}\Big\{\argmax_{j \in [K]} \mathbf{u}_j(\mathbf{x}) = c\Big\} p(\boldsymbol{\epsilon}) d\boldsymbol{\epsilon}
\end{split}
\label{eq:latent_variable_model}
\end{equation}

This generative process follows prior work in the econometrics, noisy labels and Gaussian Processes literature~\cite{train2009discrete,kendall2017uncertainties,collier2020analysis,hernandez2014mind,williams2006gaussian}, discussed further in \S \ref{sec:related_work}. First note that if we choose each stochastic component to be distributed standard Gumbel independently, $\epsilon_j \sim$ i.i.d.\ $\mathit{G}(0,1) \ \forall j$, then the predictive probabilities $p_c$ have a closed form solution that is precisely the popular softmax cross-entropy model used in training neural network classification models~\cite{train2009discrete,collier2020analysis}: 

\begin{equation}
\begin{split}
    p_c &= P(\argmax_{j \in [K]} \mathbf{u}_j(\mathbf{x}) = c) \\
    &=  \frac{\exp(\mu_c)}{\sum_{j=1}^K \exp(\mu_j)} \\
    & \iff \epsilon_j \sim \ \text{i.i.d.}\ \mathit{G}(0,1) \ \ \forall j
\end{split}
\label{eq:softmax_cross_entropy}
\end{equation}

In other words, this generative process with Gumbel noise distribution is already an implicit standard assumption when training neural network classifiers. In~(\ref{eq:softmax_cross_entropy}), the independence and identical assumptions on the noise component is however too restrictive for applications with noisy labels:
\begin{enumerate}
    \item \textbf{Identical}: for a particular input $\mathbf{x}$ some classes may have more noise than others, e.g., given an Imagenet image of a dog there may be high levels of noise on various different dog breeds but we may have high confidence that elephant classes are not present. Hence we need the level of noise to vary per class.
    \item \textbf{Independence}: if one class has a high level of noise other related classes may have high/low levels of noise. In the above example there may be correlations in the noise levels between different dog breeds.
\end{enumerate}

Our method breaks \emph{both} the independence and identical assumptions by assuming that the noise term $\boldsymbol{\epsilon}(\mathbf{x})$ is distributed multivariate Normal, $\boldsymbol{\epsilon}(\mathbf{x}) \sim \mathcal{N}(\mathbf{0}, \Sigma(\mathbf{x}))$. Computing an input-dependent covariance matrix enables modelling of inter-class label noise correlations on a per image basis. We discuss more formally in Appendix \ref{app:correlations_effect} how going beyond an independent and identical noise model can lead to improved predictions in the presence of label noise. However it also raises a number of challenges;

First there is now \textit{no closed form solution} for the predictive probabilities, Eq.\ (\ref{eq:latent_variable_model}). In order to address this, we transform the computation into an expectation and approximate using Monte Carlo estimation, Eq.\ (\ref{eq:softmax_mc_approx}).

Second, notice that the Monte Carlo estimate of Eq.\ (\ref{eq:latent_variable_model}) involves computing an $\argmax$ which makes gradient based optimization infeasible. We approximate the $\argmax$ with a temperature parameterized $\softmax_{\tau}$, Eq.\ (\ref{eq:softmax_mc_approx}). 
\begin{equation}
\begin{split}
    p_c &= P(\argmax_{j \in [K]} \mathbf{u}_j(\mathbf{x}) = c) \\
    &= \mathbb{E}_{\boldsymbol{\epsilon} \sim \mathcal{N}(0, \Sigma(\mathbf{x}))} \brac{\mathds{1} \parencurly{\argmax_{j \in [K]} \mathbf{u}_j(\mathbf{x}) = c}} \\
    &= \mathbb{E}_{\boldsymbol{\epsilon} \sim \mathcal{N}(0, \Sigma(\mathbf{x}))} \brac{(\lim_{\tau \to 0} \softmax_{\tau} \mathbf{u}(\mathbf{x}))_c} \\
    &\approx \mathbb{E}_{\boldsymbol{\epsilon} \sim \mathcal{N}(0, \Sigma(\mathbf{x}))} \brac{(\softmax_{\tau} \mathbf{u}(\mathbf{x}))_c}, \ \tau > 0 \\
    &\approx \frac{1}{S} \sum_{i=1}^S (\softmax_{\tau} \mathbf{u}^{(i)}(\mathbf{x}))_c ,\ \  \mathbf{u}^{(i)}(\mathbf{x}) \sim \mathcal{N}(\boldsymbol{\mu}(\mathbf{x}), \Sigma(\mathbf{x})).
\end{split}
\label{eq:softmax_mc_approx}
\end{equation}
The notation $(\mathbf{u}(\mathbf{x}))_c$ denotes the $c^{th}$ entry of $\mathbf{u}(\mathbf{x})$ and $S$ is the number of MC samples.

In the zero temperature limit this approximation is exact, but for non-zero temperatures $\tau$ controls a bias-variance trade-off. At lower temperatures the approximation is closer to the assumed generative process but the gradient variance is higher and vice versa. In practice $\tau$ is a hyperparameter that must be selected on a validation set. This approximation is similar to the Gumbel-softmax/Concrete distribution~\cite{gumbelsoftmax2017,concrete2017}. 
A similar derivation can be given in the multilabel classification case in which a temperature parameterized sigmoid is used as a smooth approximation to the hard $\mathds{1}$ function for each class, see Appendix \ref{app:multilabel_classification}. We analyze the effect of modelling inter-class correlations by taking a Taylor series approximation to Eq.~\ref{eq:softmax_mc_approx} in Appendix \ref{app:correlations_effect}.

\paragraph{Efficient parametrization of the covariance matrix.}
$\Sigma(\mathbf{x})$ is a $K \times K$ matrix which is a function of input $\mathbf{x}$. The memory and computational resources required to compute the full $\Sigma(\mathbf{x})$ matrix are impractical for the large-scale image classification datasets used in this paper (with $K$ up to $21k$ classes). We make a low-rank approximation to $\Sigma(\mathbf{x}) ~= V(\mathbf{x}) V(\mathbf{x})^\intercal$ where $V(\mathbf{x})$ is a $K \times R$ matrix, $R << K$. To ensure the positive semi-definiteness of the covariance matrix, we compute a K dimensional vector $d^2(\mathbf{x})$ which we add to the diagonal of $V(\mathbf{x}) V(\mathbf{x})^\intercal$. In order to sample from our noise distribution we first sample $\boldsymbol{\epsilon}_K \sim \mathcal{N}(0_K, I_{K\times K})$,
$\boldsymbol{\epsilon}_R \sim \mathcal{N}(0_R, I_{R\times R})$, then $\boldsymbol{\epsilon} = d(\mathbf{x}) \odot \boldsymbol{\epsilon}_K + V(\mathbf{x}) \boldsymbol{\epsilon}_R$, where $\odot$ denotes element-wise multiplication.

In practice we typically compute $V(\mathbf{x})$ as an affine transformation of a shared representation of $\mathbf{x}$ computed by a deep neural network. Suppose that the dimension of this representation is $D$, then the number of parameters required to compute $V(\mathbf{x})$ is $\mathcal{O}(DKR)$. For some datasets with many classes this is still impractically large. For example Imagenet-$21k$ has 21,843 classes and in the below experiments we use $R = 50$ and a ResNet-152 which has a final layer representation with $D = 2048$. So computing $V(\mathbf{x})$ requires over $2.2B$ parameters, which dwarfs the total number of parameters in the rest of the network.

In order to further reduce the parameter and computational requirements of our method we introduce a \emph{parameter-efficient} version which we use for datasets where the number of classes is too large (Imagenet-$21k$ and JFT). We parameterize $V(\mathbf{x}) = v(\mathbf{x}) \mathbf{1}_R^\intercal \odot V$ where $v(\mathbf{x})$ is a vector of dimension $R$, $\mathbf{1}_R$ is a vector of ones of dimension $R$ and $V$ is a $K \times R$ matrix of learnable parameters which is not a function of $\mathbf{x}$. Sampling the correlated noise component can be simplified, $V(\mathbf{x}) \boldsymbol{\epsilon}_R = (v(\mathbf{x}) \mathbf{1}_K^\intercal \odot V) \boldsymbol{\epsilon}_R = v(\mathbf{x}) \odot (V \boldsymbol{\epsilon}_R)$. The total parameter count of this parameter-efficient version is $\mathcal{O}(DK + KR)$ which typically reduces the memory and computational requirements dramatically for large-scale image classification datasets. For example, for Imagenet-$21k$ the number of parameters required to compute $V(\mathbf{x})$ is ~$44.8M$, a ~50$\times$ reduction. See Algorithm \ref{algo:compute_pc} for a full specification of our method.

\begin{algorithm}[t]
\SetAlgoLined
 Input: Boolean \textit{is-parameter-efficient} = \{ true/false\}\;

 compute shared representation $\mathbf{r}(\mathbf{x}) := f^{\theta}(\mathbf{x})$\;
 
 compute mean parameter $\boldsymbol{\mu}(\mathbf{x}) := W_\mu \mathbf{r}(\mathbf{x}) + \mathbf{b}_\mu$\;
 
 compute diagonal correction $\mathbf{d}(\mathbf{x}) := W_d \mathbf{r}(\mathbf{x}) + \mathbf{b}_d$;
 
 \
 
 generate $S$ standard normal samples $\boldsymbol{\epsilon}_K \sim \mathcal{N}(0_K, I_{K\times K})$, $\boldsymbol{\epsilon}_R \sim \mathcal{N}(0_R, I_{R\times R})$\;

  \eIf{is-parameter-efficient}{
   compute heteroscedastic low-rank component $v(\mathbf{x}) := W_v \mathbf{r}(\mathbf{x}) + b_v$; \\
   load homoscedastic low-rank component $V$\;
   $U(\mathbf{x}) := \boldsymbol{\mu}(\mathbf{x}) + \mathbf{d}(\mathbf{x}) \odot \boldsymbol{\epsilon}_K + v(\mathbf{x}) \odot V \boldsymbol{\epsilon}_R$\;
   }{
   compute low-rank parameters $V(\mathbf{x}) = W_V \mathbf{r}(\mathbf{x}) + b_V$\;
   $V(\mathbf{x}) := \texttt{reshape}(V(\mathbf{x}), ~[K, R])$\;
   $U(\mathbf{x}) := \boldsymbol{\mu}(\mathbf{x}) + \mathbf{d}(\mathbf{x}) \odot \boldsymbol{\epsilon}_K + V(\mathbf{x}) \boldsymbol{\epsilon}_R$\;
  }
  
  $p_c =\texttt{mean}(\softmax_{\tau} U(\mathbf{x}), axis=1)[c]$
 \caption{Computing $p_c$}
 \label{algo:compute_pc}
\end{algorithm}

\begin{table*}[tbh]
\centering
\begin{tabular}{lcccccc}
\toprule
Method & \multicolumn{2}{c}{Top-1 Acc} & \multicolumn{2}{c}{Top-5 Acc} & \multicolumn{2}{c}{NLL} \\
& 90 epochs & 270 epochs & 90 epochs & 270 epochs & 90 epochs & 270 epochs \\
\midrule
Homoscedastic & 76.5$^\dagger$ ($\pm 0.22$) & 76.7$^\dagger$ ($\pm 0.13$) & 93.0$^\dagger$ ($\pm 0.16$) & 92.9$^\dagger$ ($\pm 0.12$) & 0.98$^\dagger$ ($\pm 0.010$) & 1.08$^\dagger$ ($\pm 0.007$) \\
Het.\ Diag~\cite{collier2020analysis} & 78.3$^\dagger$ ($\pm 0.06$) & 78.7$^\dagger$   ($\pm 0.09$) & 94.0$^\dagger$ ($\pm 0.06$) & 94.0$^\dagger$ ($\pm 0.08$) & 0.88$^\dagger$ ($\pm 0.003$) & 0.95$^\dagger$ ($\pm 0.008$)\\ 
Het.\ Full (ours) & \textbf{78.5} ($\pm 0.06$) & \textbf{79.3}  ($\pm 0.10$) & \textbf{94.3} ($\pm 0.03$) & \textbf{94.5} ($\pm 0.11$) & \textbf{0.86} ($\pm 0.002$) & \textbf{0.92} ($\pm 0.006$)\\ 
\bottomrule
\end{tabular}
\caption{Results of ResNet-152 trained on ILSVRC12. For heteroscedastic models $\tau^\ast = 0.9$. Top-1 and top-5 accuracy and negative log-likelihood $\pm$ 1 standard deviation is reported. 5 runs from different random seeds are used. $^\dagger$ p < 0.05.} 
\label{table:het_vs_hom_imagenet}
\end{table*}

\section{Related work}
\label{sec:related_work}

\subsection{Heteroscedastic modelling}

\paragraph{Heteroscedastic regression.}
Heteroscedastic regression is common in the Gaussian Processes~\cite{williams2006gaussian} and econometrics literature~\cite{train2009discrete}.~\citet{bishop1997regression} introduced a heteroscedastic regression model where a neural network outputs the mean and variance parameters of a Gaussian likelihood: $y \sim \mathcal{N}(\mu(\mathbf{x}), \sigma(\mathbf{x})^2)$. The negative log-likelihood of the model is particularly amenable to interpretation, Eq.\ (\ref{eq:hetero_regression_kendall}).
\begin{equation}
    \frac{1}{N} \sum_{i=1}^{N} \frac{1}{2 \sigma(\mathbf{x}_i)^2} (y_i - \mu(\mathbf{x}_i))^2 + \frac{1}{2} \log \sigma(\mathbf{x}_i)^2.
\label{eq:hetero_regression_kendall}
\end{equation}
We see that the squared error loss term for each example is weighted inversely to the predicted variance for that example, downweighting the importance of that example's label and providing robustness to noisy labels. This heteroscedastic regression model has recently been applied to pixel-wise depth regression~\cite{kendall2017uncertainties} and in deep ensembles~\cite{lakshminarayanan2017simple}.

\paragraph{Heteroscedastic classification - diagonal covariance.}
\citet{kendall2017uncertainties} extend the heteroscedastic regression model to classification by placing a multivariate Normal distribution with diagonal covariance matrix over the softmax logits in a neural network classifier. They find that the method improves performance on image segmentation datasets which have noisy labels at object boundaries.

Closest to our methodology is the approach developed by~\citet{collier2020analysis} which we next describe. The authors reinterpret the method of~\citet{kendall2017uncertainties} as an instance of the generative framework we follow in Eq.\ (\ref{eq:latent_variable_model}). They show that this connects the method to the discrete choice modelling econometrics literature where the temperature parameterized softmax smoothing function is known as the logit-smoothed accept-reject simulator~\cite{train2009discrete,mcfadden1989method,bolduc1996multinomial}. The authors demonstrate that the softmax temperature does indeed control a bias-variance trade-off and that tuning the temperature results in different training dynamics, qualitatively improved predictions and significantly improved performance on image classification and image segmentation tasks. Both~\citet{kendall2017uncertainties} and~\citet{collier2020analysis} always use a diagonal covariance matrix for the latent distribution.

The latent variable approach to heteroscedastic classification is also standard in the Gaussian Processes literature~\cite{hernandez2014mind,williams2006gaussian}. A diagonal covariance matrix is used and a GP prior is placed on mean and log variance parameters. Again exact inference on the likelihood is intractable and different approximate inference methods are used~\cite{hernandez2014mind}.

\paragraph{Heteroscedastic segmentation - full covariance.}
\citet{monteiro2020stochastic} introduce Stochastic Segmentation Networks, a method for modelling spatially correlated label noise in image segmentation. Similar to~\citet{kendall2017uncertainties} they place a multivariate Normal distribution over the softmax logits in an image segmentation network, but share a low-rank approximation to the full covariance matrix across all the pixels in the image, capturing spatially correlated noise. Unlike our method, Stochastic Segmentation Networks are a) only applied to medical image segmentation datasets, b) do not recognise the softmax as a temperature parameterized smoothing function w.r.t.\ an assumed generative process and therefore always implicitly use a softmax temperature of 1.0 and c) do not have a parameter-efficient version of the method to enable scaling to large output vocabularies.

\paragraph{Discussion.}
Our method combines the best of this prior work and enables scaling to large-scale image classification datasets. We follow~\citet{collier2020analysis} in assuming the generative process in Eq.\ (\ref{eq:latent_variable_model}). This gives the benefits of connecting the work to the existing discrete choice modelling econometrics literature and theory. However unlike~\citet{collier2020analysis} we use a low-rank approximation to a full (non-diagonal) covariance matrix in the latent distribution. In the below experiments, we demonstrate that our combination of (a) recognizing the importance of the softmax temperature in controlling a bias-variance trade-off and (b) modelling the inter-class noise correlations yields significantly improved performance compared with individually using (a) or (b). Our parameter-efficient method also enables scaling up correlated heteroscedastic noise models to a scale unprecedented by previous work e.g.\ Imagenet-$21k$ and JFT.

\subsection{Noisy labels}

We provide a brief overview of some recent methods for training with noisy labels. Bootstrapping~\cite{reed6596training} sets the target label to be a linear combination of the ground truth label and the current model's predictions. MentorNet~\cite{MentorNet.2018} uses an auxiliary neural network, the MentorNet to output a scalar weighting for each potentially noisy example. MentorMix~\cite{jiang2020beyond} adds \textit{mixup} regularization~\cite{zhang2018mixup} to the MentorNet approach. Co-teaching~\cite{CoTeaching.2018} jointly trains two neural networks. Each network makes predictions on a mini-batch and the small loss samples are then fed to the other network for learning. It is assumed that small loss examples are more likely to have clean labels.~\citet{cao2020heteroskedastic} propose a heteroscedastic adaptive regularization scheme which increases the regularization strength on high noise examples.

\begin{table*}[tbh]
\centering
\begin{tabular}{lcccccc}
\toprule
Method &
  \multicolumn{3}{c}{Webvision} &
  \multicolumn{3}{c}{ILSVRC12} \\
 & Top-1 Acc & Top-5 Acc & NLL & Top-1 Acc & Top-5 Acc & NLL \\
\midrule
\citet{lee2018cleannet} & 69.1 & 86.7 & - & 61.0 & 82.0 & - \\
\citet{MentorNet.2018} & 72.6 & 88.9 & - & 64.2 & 84.8 & - \\
\citet{guo2018curriculumnet} & 72.1 & 89.2 & - & 64.8 & 84.9 & - \\
\citet{saxena2019data} & 67.5 & - & - & - & - & - \\
\citet{jiang2020beyond} & 74.3 & 90.5 & - & 67.5 & 87.2 & - \\
\citet{cao2020heteroskedastic} & 75.0 & 90.6 & - & 67.1 & 86.7 & - \\
\midrule
Homoscedastic & 76.1$^\dagger$ ($\pm 0.07$) & 91.2$^\dagger$ ($\pm 0.07$) & 1.03$^\dagger$ ($\pm 0.006$) & 67.0$^\dagger$ ($\pm 0.08$) & 86.0$^\dagger$ ($\pm 0.11$) & 1.49$^\dagger$ ($\pm 0.008$) \\
Het.\ Diag~\cite{collier2020analysis} & 76.2$^\dagger$ ($\pm 0.15$) & 91.4$^\dagger$ ($\pm 0.08$) & 1.01$^\dagger$ ($\pm 0.002$) & 67.3$^\dagger$ ($\pm 0.10$) & 86.1$^\dagger$ ($\pm 0.07$) & 1.47$^\dagger$ ($\pm 0.004$)\\ 
Het.\ Full (ours) & \textbf{76.6} ($\pm 0.13$) & \textbf{92.1} ($\pm 0.09$) & \textbf{0.98} ($\pm 0.004$) & \textbf{68.6} ($\pm 0.17$) & \textbf{87.1} ($\pm 0.13$) & \textbf{1.41} ($\pm 0.010$)\\ 
\bottomrule
\end{tabular}
\caption{WebVision 1.0 results. For het.\ models, $\tau^\ast = 0.9$. Top-1 and top-5 accuracy and negative log-likelihood $\pm 1$ standard deviation is reported for both the WebVision and ILSVRC12 validation sets. 5 runs from different random seeds are used for the homo/heteroscedastic methods, all other results are taken from the literature. $^\dagger$ p-value < 0.05.}
\label{table:webvision}
\end{table*}

\section{Experiments}

Our main experiment is to evaluate our method on four large-scale image classification datasets and show significant performance improvements over baseline methods. We also provide qualitative analysis of the learned covariance matrices in \S \ref{sec:qualitative}. We analyse the effect of our method on the representations learned by the network in \S \ref{sec:rep_learning}. Finally, in \S \ref{sec:deep_ensembles} we combine our method with Deep Ensembles~\cite{lakshminarayanan2017simple} to yield a method with full predictive uncertainty.

We provide code which implements our method as a TensorFlow Keras layer~\cite{abadi2016tensorflow,chollet2015keras}, in the supplementary material. The layer is a drop-in replacement for the final layer of a classifier, requiring only a simple one line code change from:
\begin{lstlisting}[language=Python]
logits = tf.keras.layers.Dense(...)(x)
\end{lstlisting}
to
\begin{lstlisting}[language=Python]
logits = MCSoftmaxDenseFA(...)(x).logits
\end{lstlisting}
No other changes, including to the loss function are required.

\paragraph{Imagenet ILSVRC12.} Table \ref{table:het_vs_hom_imagenet} shows the results on Imagenet ILSVRC12, a dataset of over 1.2M training images with $1k$ classes. ILSVRC12 is known to have noisy labels~\cite{beyer2020we}. Only one class can be present for each image. We train a ResNet-152~\cite{he2016deep} for 90 and 270 epochs (further details in Appendix \ref{app:experimental_details}). Hyperparameters are tuned on a validation set of 50,000 examples that we split off from the training set. As is standard, we report results on the official ILSVRC12 validation set. For our method we use a softmax temperature of 0.9 and covariance matrix rank of 15.

When trained for 270 epochs our method has a validation set top-1 accuracy of 79.3\% statistically significantly better than the baseline models based on an unpaired two-tailed t-test. We compare to a homoscedastic baseline (standard neural network training), against which our method improves the top-1 accuracy by 2.6\%. Compared to the diagonal covariance method we see a smaller improvement of 0.6\% for top-1 accuracy, suggesting that much of the gain is from the diagonal covariance matrix entries, but that the off-diagonal terms give further improvements. We note that we are the first to evaluate the heteroscedastic diagonal model on ILSVRC12. A sensitivity analysis to the number of MC samples is provided in Appendix~\ref{app:mc_samples_sensitivity}. An ablation study equalizing the number of parameters in the heteroscedastic and homoscedastic models is provided in Appendix~\ref{app:equal_params}.

Prior work has shown that neural networks fit cleanly labelled data points first and then fit examples with noisy labels~\cite{krueger2017closer}. The purpose of the 90 epoch ablation is to demonstrate the heteroscedastic models gain more from longer training schedules than the homoscedastic model. By overfitting less to noisy labels the heteroscedastic models can be trained for longer e.g., only our heteroscedastic sees improved top-5 accuracy from training for 270 epochs while the increase in top-1 accuracy from the longer training schedule increases from 0.2\% from the homoscedastic model, to 0.4\% for the diagonal covariance heteroscedastic model to 0.8\% for our model. This demonstrates that despite the parameter count of the heteroscedastic models being higher than the homoscedastic models, the additional parameters do not lead to more overfitting, see Appendix~\ref{app:equal_params}.

\begin{table*}[tbh]
\centering
\begin{tabular}{lcccc}
\toprule
Method & Imagenet-$21k$ gAP & Imagenet-$21k$ NLL & JFT gAP & JFT NLL \\
\midrule
Homoscedastic & 45.9$^\dagger$ ($\pm 0.14$) & 3.65$^\dagger$ ($\pm 0.010$) & 63.1$^\dagger$ ($\pm 0.22$) & 20.12$^\dagger$ ($\pm 0.090$) \\
Heteroscedastic Diag $\tau = 1.0$~\cite{collier2020analysis} & 45.9$^\dagger$   ($\pm 0.06$) & 3.64$^\dagger$ ($\pm 0.003$) & 63.1$^\dagger$ ($\pm 0.03$)  & 19.95$^\dagger$ ($\pm 0.022$)\\ 
Heteroscedastic Diag $\tau^* = 0.15$~\cite{collier2020analysis} & 46.8$^\dagger$   ($\pm 0.04$) & 3.63 ($\pm 0.002$) & 64.1$^\dagger$ ($\pm 0.07$) & 19.61$^\dagger$ ($\pm 0.030$)\\ 
Heteroscedastic PE $\tau^* = 0.15$ (ours) & \textbf{47.0}  ($\pm 0.08$) & \textbf{3.62} ($\pm 0.005$) & \textbf{64.7} ($\pm 0.06$) & \textbf{19.34} ($\pm 0.026$)\\ 
\bottomrule
\end{tabular}
\caption{Imagenet-$21k$ and JFT results for heteroscedastic and homoscedastic models. Heteroscedastic PE, is the parameter-efficient version of our method. The test set global average prevision (gAP) and negative log-likelihood $\pm$ 1 standard deviation is reported. 5 runs from different random seeds are used. $^\dagger$ p-value < 0.05.}
\label{table:het_vs_hom_imagenet$21k$_jft}
\end{table*}

\paragraph{WebVision.} WebVision 1.0~\cite{li2017webvision} is a popular benchmark for noisy label techniques with the same 1,000 classes as Imagenet ILSVRC12. Labels are gathered through a noisy automated process based on co-occuring text with the images which are scraped from the Web. The training set consists of 2.4M examples and we report results on the validation set, as is standard, which has 50,000 examples.

Following other approaches in the literature~\cite{MentorNet.2018,guo2018curriculumnet,jiang2020beyond,cao2020heteroskedastic} we use a InceptionResNet-v2 architecture~\cite{inceptionresnet2017} with the same softmax temperature and covariance matrix rank as the above ILSVRC12 experiments. Other hyperparameters are taken from~\citet{jiang2020beyond}, however we adopt a longer training schedule of 95 epochs. Following previous methods we report top-1 and top-5 accuracy on both the WebVision validation set and on the ILSVRC12 validation set. For the methods we implement we also report negative log-likelihood.

Our method achieves a new state-of-the-art on WebVision 1.0 with top-1 accuracy of 76.6\% and top-5 accuracy of 92.1\% on the WebVision validation set. See Table \ref{table:webvision} for detailed results. Our method improves upon the best previously published WebVision top-1 accuracy by 1.6\% and by 1.1\% over the best published ILSVRC12 top-1 accuracy, when training on WebVision. Our baseline homoscedastic and diagonal heteroscedastic methods are strong relative to the previously published WebVision results, perhaps due to our longer than typical 95 epoch training schedule. We note however there is no standard training schedule for WebVision and our experiments showed that none of the models converge with shorter training schedules.

\paragraph{Multilabel datasets.} We can also apply our method to multilabel datasets which may have more than one class in each image. The same latent variable formulation can be used but with temperature parameterized sigmoid smoothing function (see Appendix \ref{app:multilabel_classification}). Imagenet-$21k$ and JFT are two large-scale multilabel image classification datasets.

Imagenet-$21k$ is a larger version of the standard ILSVRC-2012 Imagenet benchmark~\cite{deng2009imagenet,kolesnikov2019big,beyer2020we,collier2020analysis}. It has over 12.8 million training images with 21,843 classes. No standard train/test split is provided, so we use 4\% of the dataset as a validation set and a further 4\% as the test set.

JFT-300M~\cite{distillation2015,chollet2017xception,sun2017revisiting,kolesnikov2019big} is a dataset introduced by~\citet{distillation2015} with over 300M training images and validation and test sets with 50,000 images. JFT has over 17k classes and each image can have more than one class (average 1.89 per image). The labels were collected automatically, with 20\% of them estimated to be noisy~\cite{sun2017revisiting}.

For Imagenet-$21k$ we train a Resnet-152~\cite{he2016deep} for 90 epochs. Whereas for JFT we train a Resnet-50~\cite{he2016deep} for 30 epochs. Further experimental setup details in Appendix \ref{app:experimental_details}. Sigmoid temperature of 0.15 is used for the heteroscedastic methods. For our method the covariance matrix rank is set to 50 and as the number of classes is large for both datasets, we use the parameter-efficient version of our method, \S \ref{sec:method}.

Table \ref{table:het_vs_hom_imagenet$21k$_jft} shows the test set results for Imagenet-$21k$ and JFT. Global average precision (gAP), the average precision over all classes, is the metric of interest. The diagonal covariance heteroscedastic method with tuned sigmoid temperature provides gains over standard neural network training and modelling the full covariance martix provides further impovements in gAP for both datasets. On Imagenet-$21k$ the additional improvement from modelling the full covariance matrix is more marginal than for JFT. We include an ablation to demonstrate the importance of the smoothing function temperature parameter. For both datasets, the performance of the diagonal method is significantly degraded at $\tau = 1.0$, which corresponds to the model of~\citet{kendall2017uncertainties}, compared to $\tau^* = 0.15$. We note that~\citet{collier2020analysis} have already empirically demonstrated the temperature controls a bias-variance trade-off in the training dynamics of the diagonal covariance heteroscedastic model.

\begin{table}[tbh]
\centering
\begin{tabularx}{\linewidth}{LLc}
\toprule
Class A & Class B & Avg.\ Cov.\ \\
\midrule
partridge & ruffed grouse, partridge, Bonasa umbellus & -0.46 \\
\hline
projectile, missile & missile & -0.44 \\
\hline
maillot, tank suit & maillot & -0.42 \\
\hline
screen, CRT screen & monitor & -0.37 \\
\hline
tape player & cassette player & -0.34 \\
\hline
Welsh springer spaniel & Blenheim spaniel & 0.28 \\
\hline
standard schnauzer & miniature schnauzer & 0.27 \\
\hline
French bulldog & Boston bull, Boston terrier & 0.27 \\
\hline
EntleBucher & standard schnauzer & 0.26 \\
\hline
tennis ball & racket, racquet & -0.26 \\
\bottomrule
\end{tabularx}
\caption{Class pairs with the top-10 absolute covariance, averaged over the ILSVRC12 validation set.}
\label{table:top_average_cov}
\end{table}

\newcommand{\imagenet}[4] {
  \includegraphics[width=1.93cm]{images/imagenet_examples/imagenet_image_#1.pdf} &
  \vspace{-1.9cm}\begin{tabular}[t]{p{5cm}} Het pred: \small{\textsf{#2}} \\ Hom pred: \small{\textsf{#3}} \\ Covariance: #4\end{tabular}}
  
\begin{table}[tbh]
\centering
\begin{tabularx}{\linewidth}{p{2cm}p{5cm}}
\toprule
Image & Predictions \& covariance\\
\midrule
\imagenet{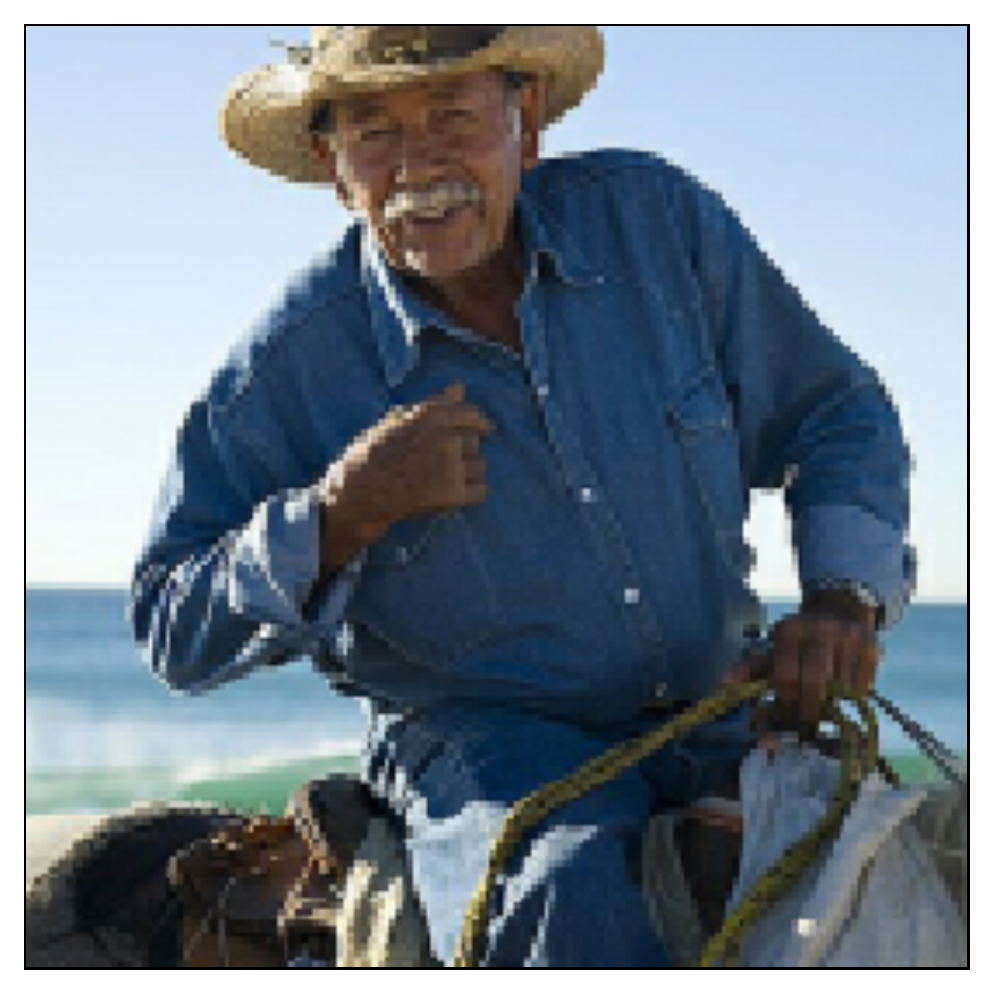}{cowboy hat}{ten-gallon hat, cowboy boot}{-4.26} \\
\hline
\imagenet{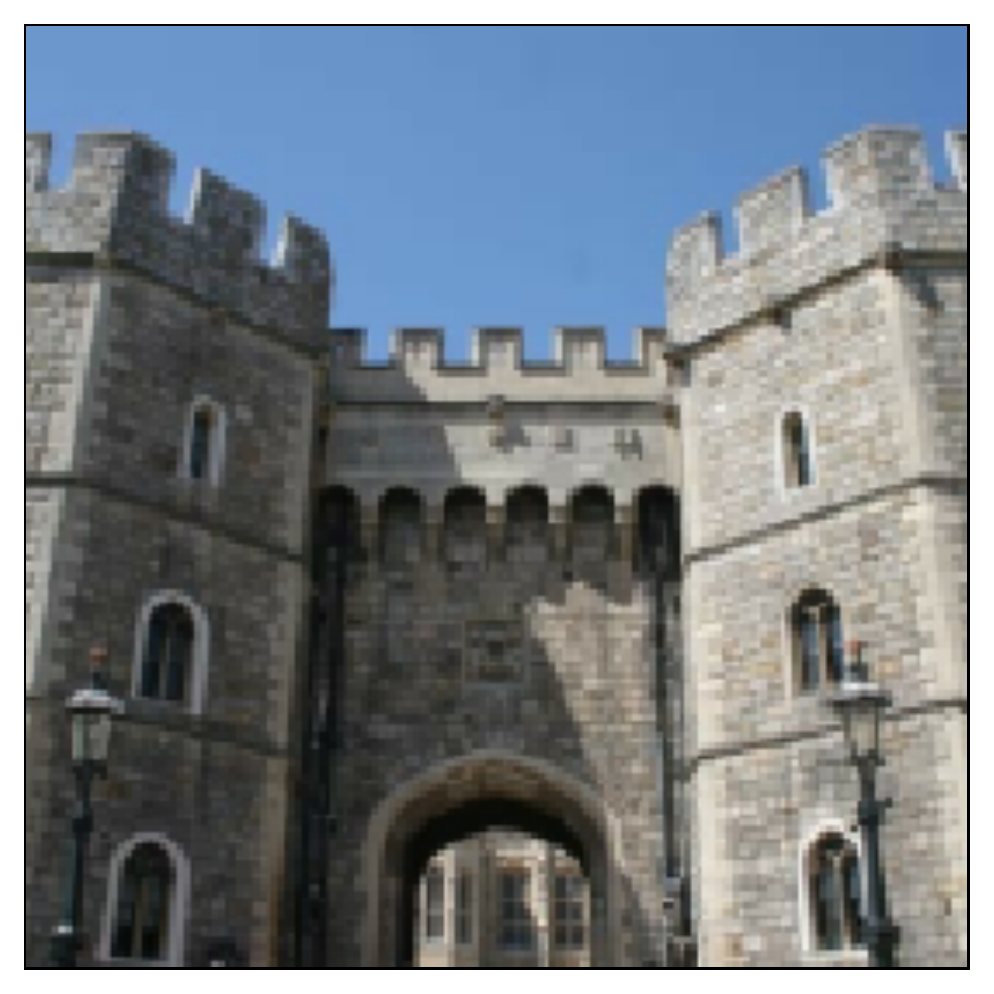}{castle}{palace}{-1.17 } \\
\hline
\imagenet{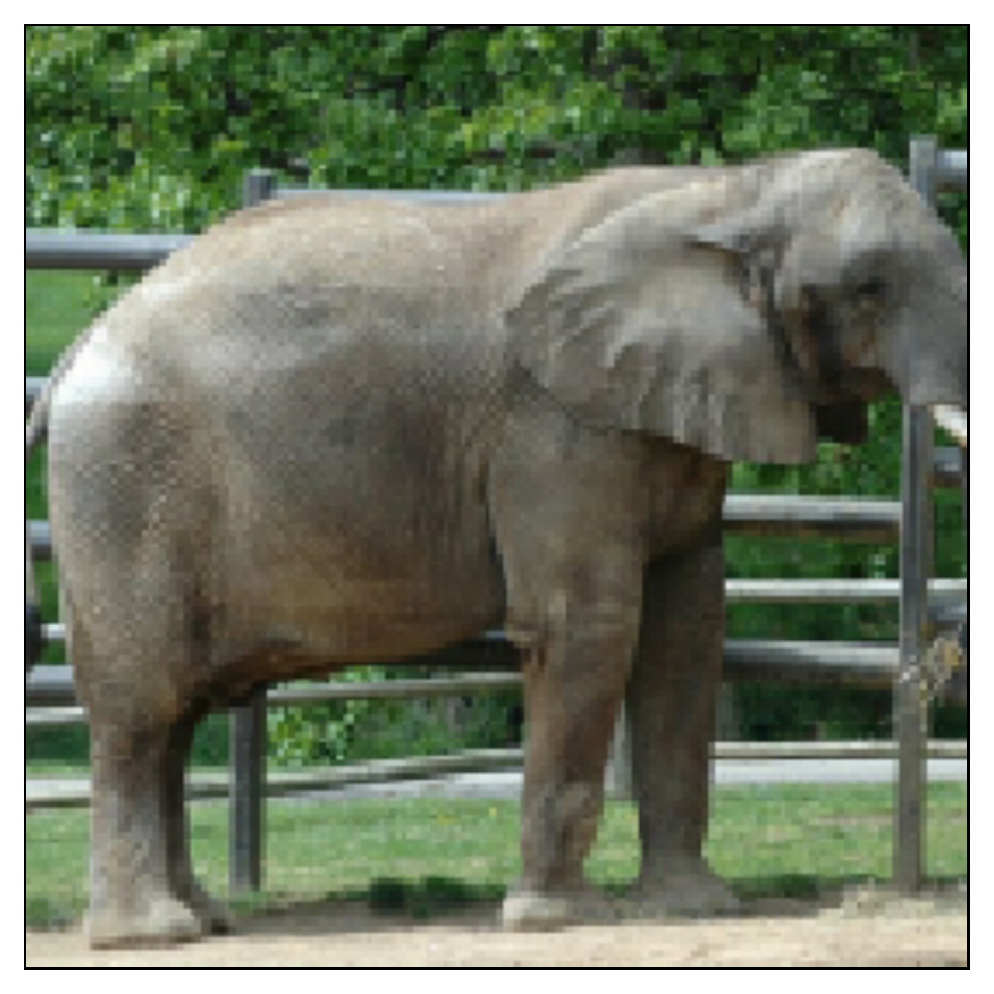}{African elephant, Loxodonta africana}{Indian elephant, Elephas maximus}{0.78 } \\
\hline
\imagenet{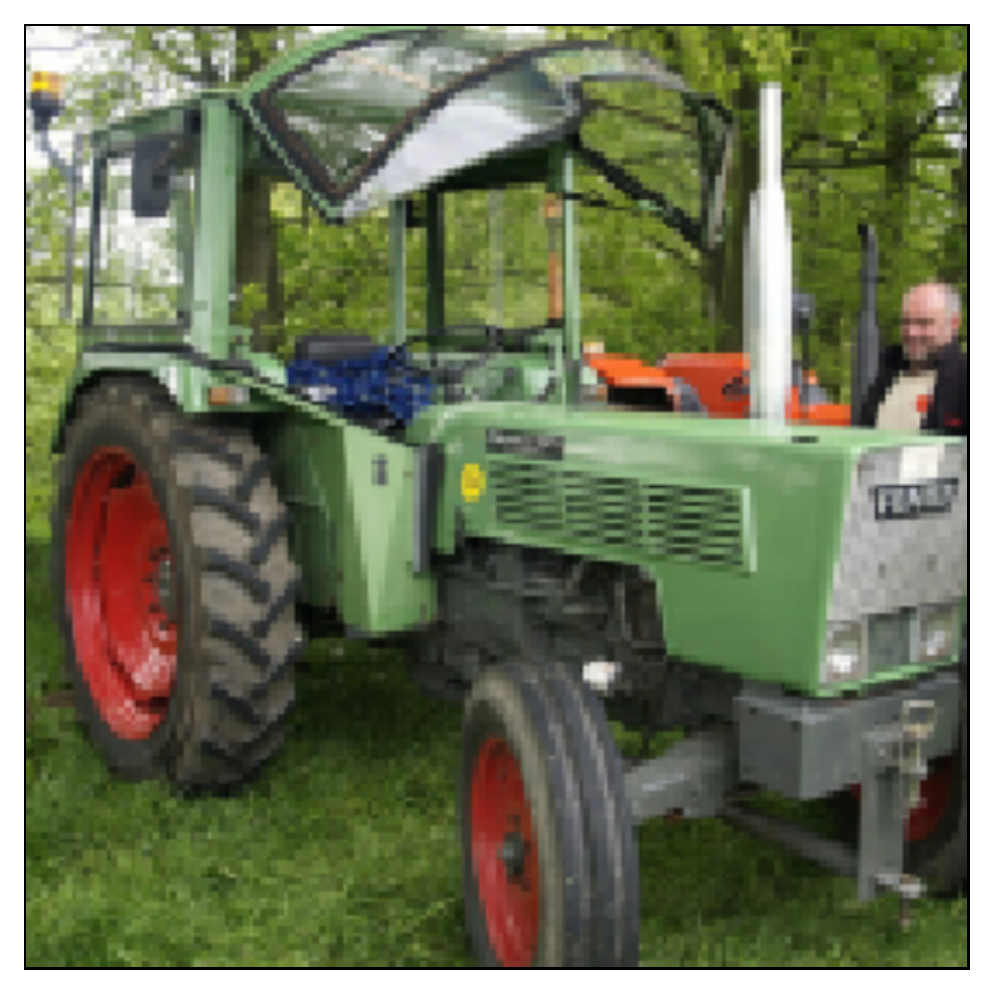}{tractor}{harvester, reaper}{1.37 } \\
\hline
\imagenet{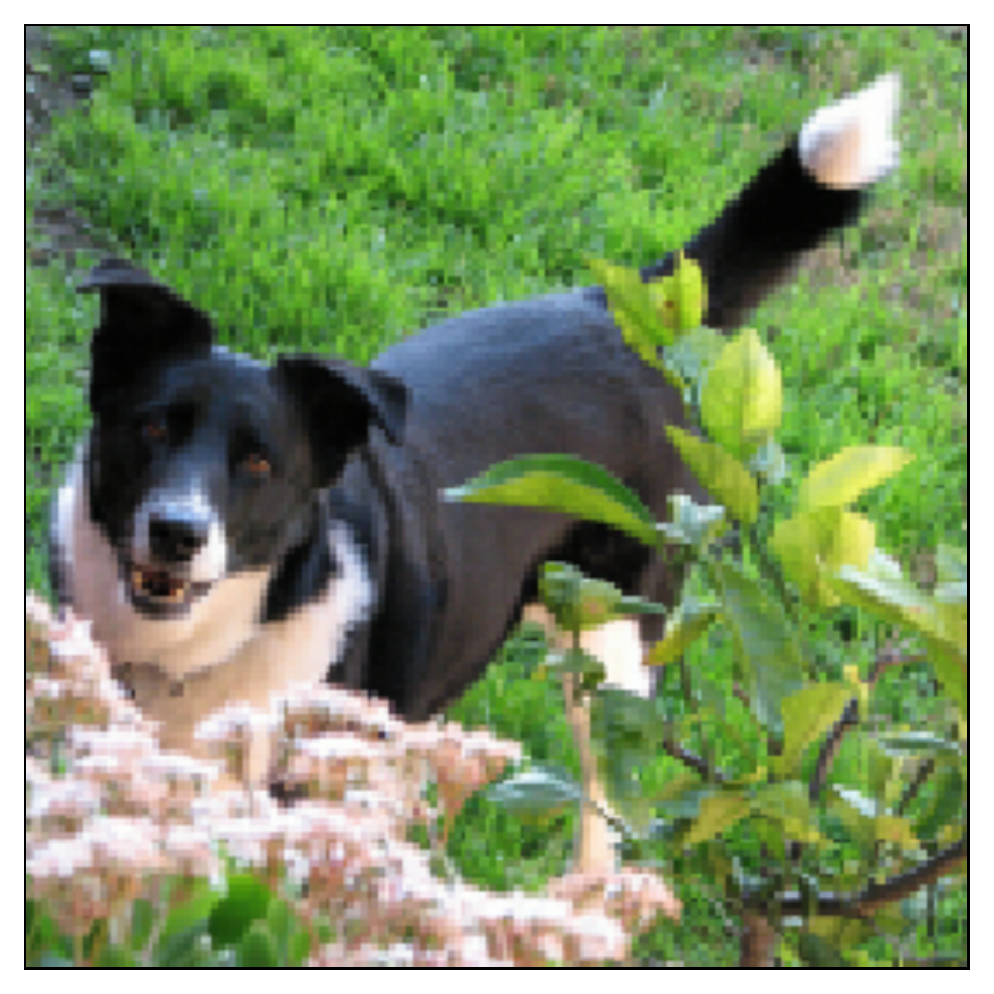}{Border collie}{Cardigan, Cardigan Welsh corgi}{1.55 } \\
\hline
\imagenet{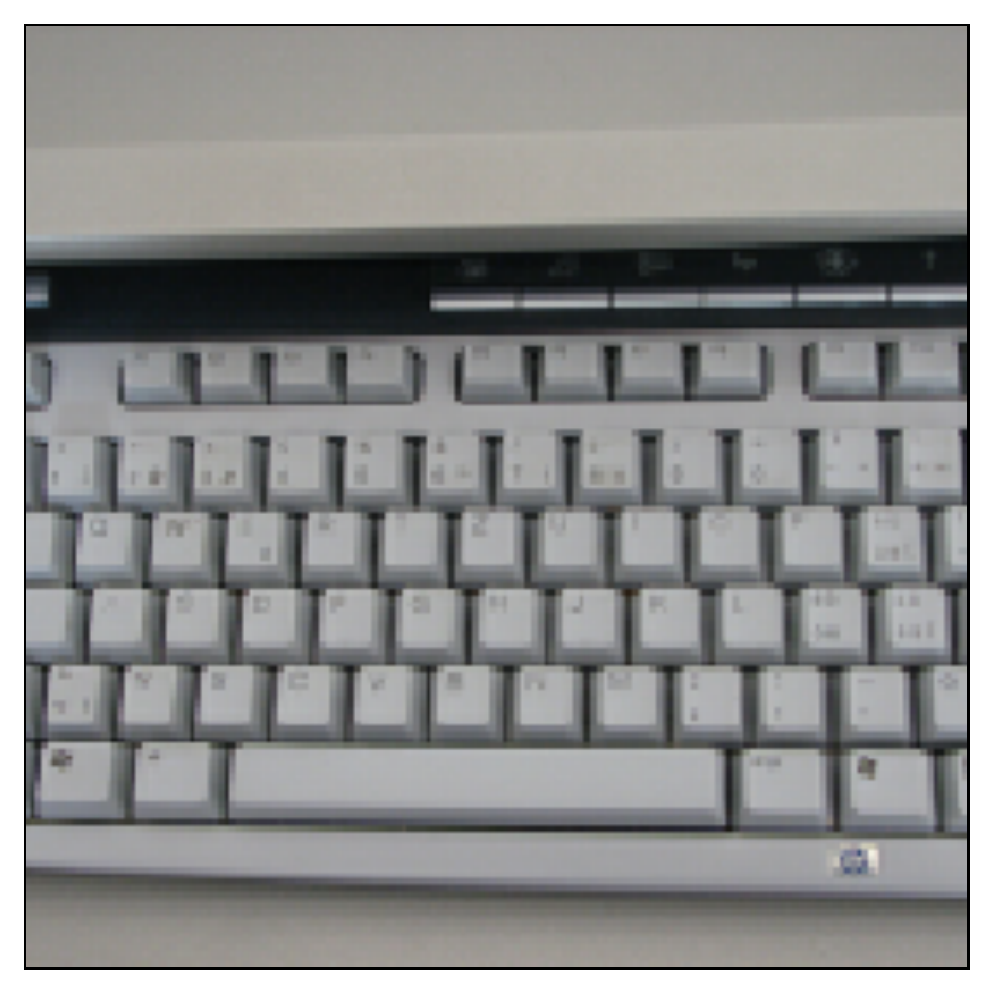}{computer keyboard, keypad}{space bar}{1.82 } \\
\hline
\imagenet{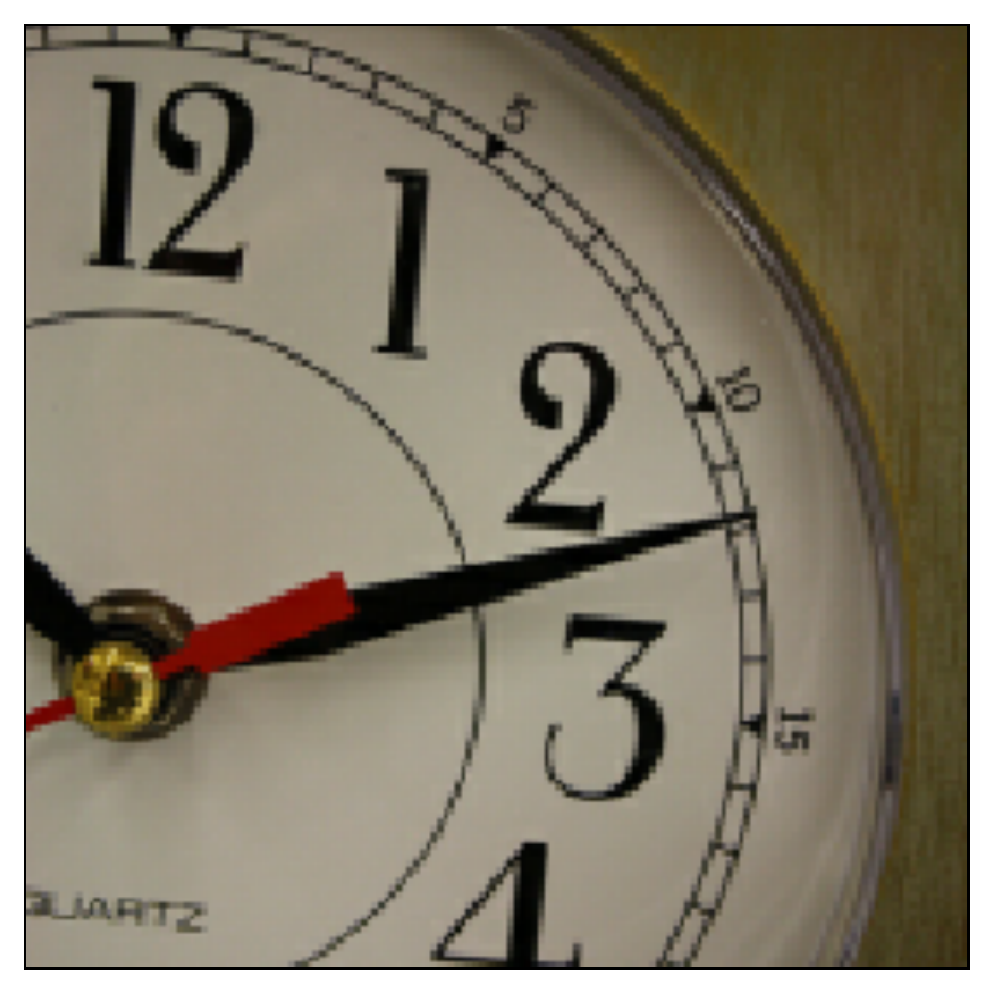}{analog clock}{wall clock}{2.46 } \\
\hline
\imagenet{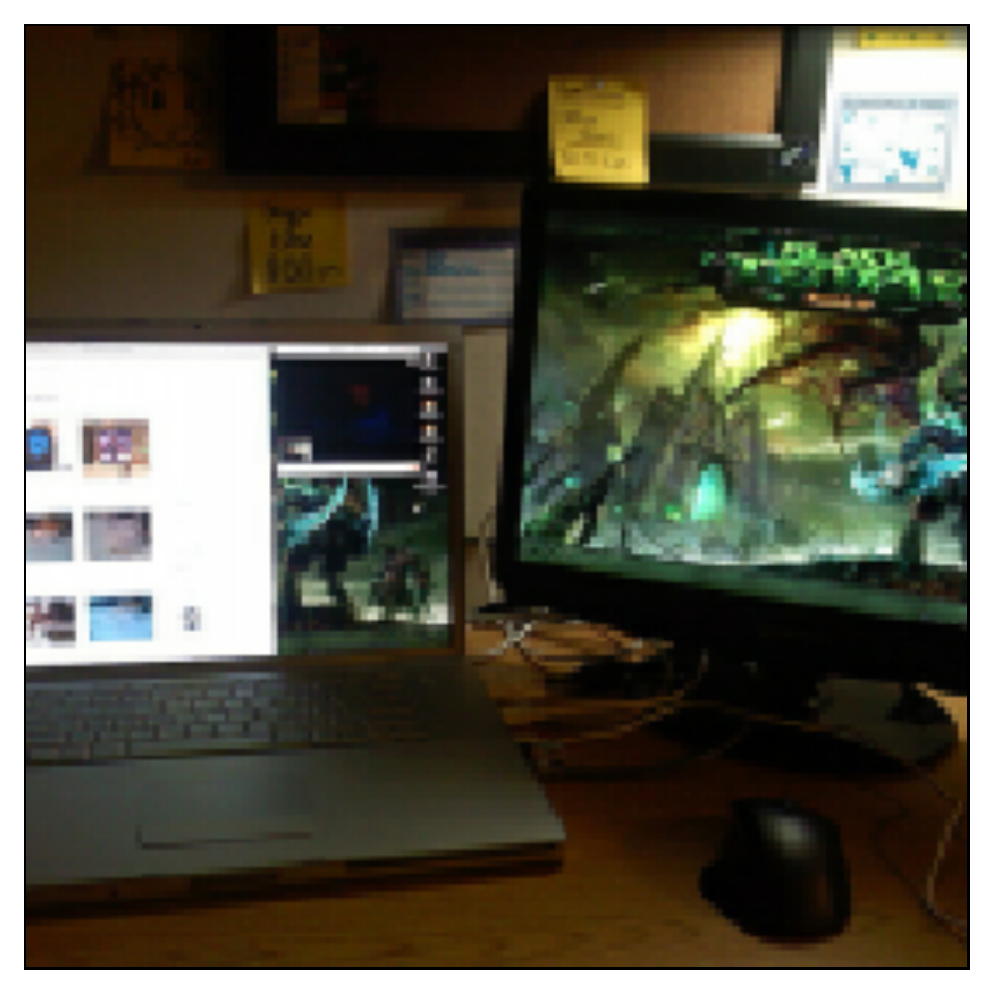}{monitor}{desktop computer}{3.07 } \\
\bottomrule
\end{tabularx}
\caption{8 ILSVRC12 test set examples where the heteroscedastic model is correct, the homoscedastic model is incorrect and the absolute covariance between the heteroscedastic and homoscedastic prediction is the largest of all classes to the heteroscedastic prediction.\vspace{-0.7cm}}
\label{table:top_cov}
\end{table}


\subsection{Qualitative analysis of learned covariances}
\label{sec:qualitative}

We can examine cases where our heteroscedastic model makes the correct prediction but the standard homoscedastic model is incorrect. Particularly we are interested in understanding whether there is structure in the covariance matrix of our method which helps explain the correct prediction. Note that we can reconstruct the full covariance matrix as $\Sigma(\mathbf{x}) = diag(\mathbf{d}(\mathbf{x})^2) +  V(\mathbf{x}) V(\mathbf{x})^\intercal$.

In Table \ref{table:top_cov}, we list 8 cases when the homoscedastic prediction has the highest absolute covariance with the correct class and the heteroscedastic prediction is correct but the homoscedastic prediction is incorrect. We see that these cases broadly fall in two categories which can easily contribute to noisy labels; 1) \textbf{substitutes}, e.g., a castle and a palace are two easily confused classes and 2) \textbf{co-occurrence}, e.g., a computer keyboard and a space bar are likely to occur in the same image but which one is considered the most prominent class by the annotator may be unclear.

In Table \ref{table:top_average_cov} we look at the class pairs with the highest absolute covariance \textit{averaged} over the ILSVRC12 validation set. We compute the average covariance matrix over the 50,000 validation set images and extract the top-10 entries by absolute value. The class pairs, all are substitutes for each other or commonly co-occur. There are just under 1M possible class pairs, it is remarkable that the average covariance matrix exhibits such clear and consistent structure.

\begin{figure}
    \centering
    \includegraphics[width=0.36\textwidth]{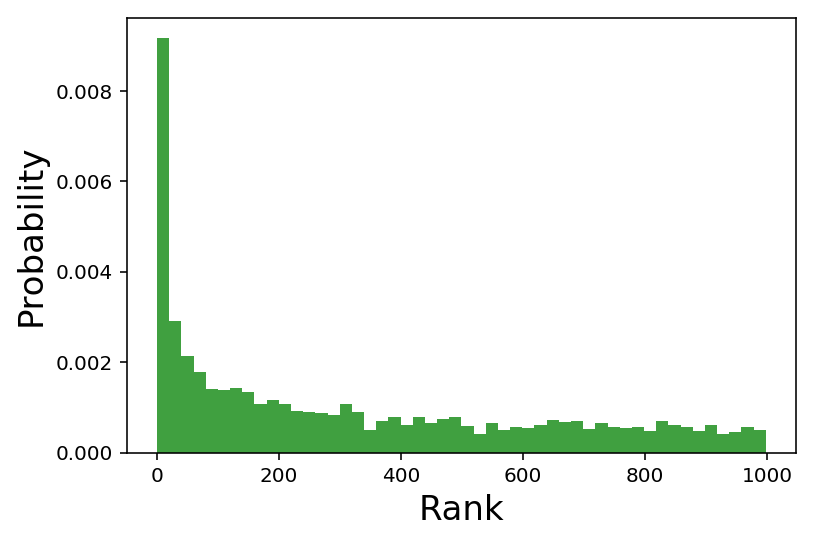}
    \caption{Histogram of the sorted rank of the absolute covariance between the ground truth class and the homoscedastic model's predicted class for ILSVRC12 validation set examples which the heteroscedastic model predicted correctly and the homoscedastic model did not.}
    \label{fig:covariance_rank}
\end{figure}

We now conduct a simple analysis to show that the above examples are not anecdotal but that the learned covariance matrices are structured. For one homoscedastic and one heteroscedastic model selected from the previous Imagenet ILSVRC12 experiments there are 3,565 out of 50,000 validation set images where our model makes a correct prediction but the homoscedastic model is incorrect. Figure (\ref{fig:covariance_rank}) shows a histogram of the rank of the sorted absolute covariance between the homoscedastic prediction and the correct class. Clearly the class the homoscedastic model has incorrectly predicted, is much more likely to have a high absolute covariance to the correct class in our \textit{learned} covariance matrix. Assuming that the incorrect homoscedastic prediction is sometimes a plausible label then we see that the heteroscedastic model has learned to associate the noise on the correct class with the noise on plausible alternatives. This is consistent with our analysis of the 2nd order Taylor series approximation to our model's log-likelihood, Appendix~\ref{app:correlations_effect}. We expect to see covariances of commonly confused class pairs to be strengthened during training.

\subsection{Transfer of learned image representations}
\label{sec:rep_learning}

Often large-scale image classification datasets are used to pre-train representations which are fine-tuned on smaller specialized datasets~\cite{zhai2020largescale,kolesnikov2019big,sun2017revisiting,djolonga2020robustness,raghu2019transfusion}. We wish to evaluate the transferability of the representations learned by our model. We hypothesize that by overfitting less to noisy labels and learning a better model of the upstream pre-training distribution, our method should learn more general representations.

We test the JFT models on the VTAB~\cite{zhai2020largescale}. VTAB consists of 19 unseen downstream classification datasets which cover a variety of visual domains and tasks. We evaluate using the \vtab protocol, where each model is fine-tuned on only 1000 datapoints for each downstream dataset. For all 19 datasets the fine-tuned model is homoscedastic, i.e., the heteroscedastic output layer is only ever used for upstream pre-training. The output layer of the network is removed and replaced with a untrained homoscedastic output layer for fine-tuning. For downstream fine-tuning we use the standard hyperparameters and data augmentation settings specified by~\citet{kolesnikov2019big}. The \vtab~score is an average of the accuracy on all 19 datasets.

Table~\ref{tab:vtab} shows \vtab~scores. Our parameter-efficient heteroscedastic model, which captures correlations in the JFT label noise, improves the \vtab~score by 0.88\% over the homoscedastic baseline and by 0.22\% over the heteroscedastic diagonal model. We stress that the downstream models are trained \textit{without} a heteroscedastic output layer, so our experiment demonstrates that a model trained upstream on JFT with a heteroscedastic output layer learns representations which transfer better than a homoscedastic or a heteroscedastic diagonal model.

\begin{table}[]
\centering
\begin{tabular}{@{}ll@{}}
\toprule
 & \vtab~score \\ \midrule
Homoscedastic & $70.46^\dagger$ $\pm$ 0.5 \\
Heteroscedastic Diag & 71.12 $\pm$ 0.19 \\
Heteroscedastic PE & \textbf{71.34} $\pm$ 0.23 \\ \bottomrule
\end{tabular}
\caption{\vtab~score $\pm$ 1 standard deviation for ResNet50 models pre-trained on JFT and fine-tuned on 19 diverse image classification datasets. $\dagger$ p-value < 0.05.}
\label{tab:vtab}
\end{table}

\subsection{Deep Ensembles for Full Predictive Uncertainty}
\label{sec:deep_ensembles}

\begin{table}[tbh]
\centering
\begin{tabular}{lccc}
\toprule
Method & Top-1 Acc & NLL & ECE \\
\midrule
Hom.\ Single Model & 76.1 & 0.943 & 0.0392 \\
Hom.\ Ensemble 4$\times$ & 77.5 & 0.877 & 0.0305\\ 
\midrule
Het.\ Single Model  & 77.5 & 0.898 & 0.033\\ 
Het.\ Ensemble 4$\times$ & \textbf{79.5} & \textbf{0.79} & \textbf{0.015}\\ 
\bottomrule
\end{tabular}
\caption{Deep Ensemble results ResNet-50 on Imagenet ILSVRC12. For heteroscedastic models, $\tau^\ast = 1.5$.}
\label{table:deep_ensembles}
\end{table}

Our method estimates heteroscedastic aleatoric uncertainty i.e., input-dependent fundamental label noise in the data. Most approaches in the Bayesian neural networks literature focus on estimating epistemic uncertainty over the network's parameters~\cite{gal2016dropout,gal2017concrete,blundell2015weight, neal1995bayesian, wilson2020bayesian, wenzel2020good,lakshminarayanan2017simple}. Our method can be easily combined with many of these methods, giving an estimate of full predictive uncertainty. We successfully combine our method with Deep Ensembles~\cite{lakshminarayanan2017simple}, a method inspired by Bayesian approaches, that compares favorably in uncertainty modelling benchmarks~\cite{snoek2019can, gustafsson2019evaluating}. To form a deep heteroscedastic ensemble we train each ensemble member with our heteroscedastic layer and average the predictions of the ensemble members, as in a standard (homoscedastic) Deep Ensemble.

A ResNet-50~\cite{he2016deep} trained on Imagenet ILSVRC12 is a standard uncertainty quantification benchmark~\cite{dusenberry2020efficient,wen2020combining}. We train our method and a standard homoscedastic method using 4 different random seeds. We then create an ensemble of each method. We use rank 15 covariance matrices and softmax temperature of 1.5 for the heteroscedastic method and train for 180 epochs. Homoscedastic models are trained for 90 epochs as this maximizes validation set log-likelihood. Table~\ref{table:deep_ensembles} shows the results. We also report expected calibration error~\cite{guo2017calibration}, a standard metric for uncertainty benchmarks which measures how calibrated a network's predictions are independent of its performance.

Going from a single homoscedastic model to a ensemble of homoscedastic models provides a substantial improvement in all metrics as does going from a single homoscedastic to our heteroscedastic model. However the best top-1 accuracy, negative log-likelihood and ECE are achieved by a Deep Ensemble of heteroscedastic models. The improvement in top-1 accuracy for the heteroscedastic ensemble (79.5\%) compared to the homoscedastic model (76.1\%) is greater than the combined gains from ensembling homoscedastic models and the heteroscedastic single model, perhaps due to additional diversity of the ensemble members. Code to reproduce these results and a leaderboard of methods is available publicly\footnote{https://github.com/google/uncertainty-baselines/tree/master/baselines/imagenet}.

\section{Conclusion}

We have introduced a new probabilistic method for deep classification under input-dependent label noise. Our method models inter-class correlations in the label noise. We show that the learned correlations correspond to known sources of label noise such as two classes being visually similar or co-occurring. The proposed method scales to very large-scale datasets and we see significant gains on Imagenet ILSVRC12, Imagenet-$21k$ and JFT. We set a new state-of-the-art top-1 accuracy on WebVision. The representations learned by our model on JFT transfer better when fine-tuned on 19 datasets from the VTAB. We combine our method with Deep Ensembles, giving a method for full predictive uncertainty estimation and see substantially improved accuracy, log-likelihood and expected calibration error on ILSVRC12.

\clearpage

{\small
\bibliography{egbib}
}

\clearpage
\appendix

\section{Multilabel classification}
\label{app:multilabel_classification}

The extension to the multilabel case is simple. Each label is treated independently conditional on the latent $\mathbf{u}$, the $\argmax$ is replaced with the $\mathds{1}$ indicator function which can be approximated with a temperature parameterized $\sigmoid$.

\begin{equation}
\begin{split}
    p_c &= P(y_c = 1 | \mathbf{x}) \\
    &= P(\mathds{1} \parencurly{u(\mathbf{x})_c > 0}) \\
    &= \mathbb{E}_{\boldsymbol{\epsilon} \sim \mathcal{N}(0, \Sigma(\mathbf{x}))} \brac{\mathds{1} \parencurly{u(\mathbf{x})_c > 0}} \\
    &= \mathbb{E}_{\boldsymbol{\epsilon} \sim \mathcal{N}(0, \Sigma(\mathbf{x}))} \brac{\lim_{\tau \to 0} \sigmoid_{\tau} u(\mathbf{x})_c} \\
    &\approx \mathbb{E}_{\boldsymbol{\epsilon} \sim \mathcal{N}(0, \Sigma(\mathbf{x}))} \brac{\sigmoid_{\tau} u(\mathbf{x})_c}, \ \tau > 0 \\
    &\approx \frac{1}{S} \sum_{i=1}^S \sigmoid_{\tau} u^{(i)}(\mathbf{x})_c ,\ \  \mathbf{u}^{(i)}(\mathbf{x}) \sim \mathcal{N}(\boldsymbol{\mu}(\mathbf{x}), \Sigma(\mathbf{x})).
\end{split}
\label{eq:sigmoid_mc_approx}
\end{equation}

\section{Experimental details}
\label{app:experimental_details}

\paragraph{Imagenet ILSVRC12.} We train a ResNet-152v2 \cite{he2016deep} for 90 and 270 epochs. Stochastic gradient descent with momentum factor $= 0.9$ and initial learning rate of $0.1$ is used as the optimizer. The learning rate is decayed by a factor of $10\times$ at 30, 60 and 80 epochs for the 90 epoch training and 90, 180 and 240 epochs for the 270 epoch training. During the first 5 epochs the learning rate follows a linear warm-up schedule. $L2$ regularization with weight $10^{-3}$ is used. Standard data augmentation is used during training, an Inception crop followed by resizing to $224 \times 224$ and left-right horizontal flipping. All inputs are scaled to the $[-1, 1]$ range. Models are trained on $4 \times 4$ Google Cloud TPUv3s with a batch size of 1,024. For all heteroscedastic models the optimal $\sigmoid$ temperature of $0.9$ is tuned on the validation set. A rank 15 low-rank approximation to the covariance matrix is used for our method. 10,000 MC samples are used at train and eval time for the heteroscedastic methods.

\paragraph{WebVision 1.0.} We train an Inception-ResNetv2 \cite{inceptionresnet2017} for 95 epochs. A distributed asynchronized stochastic gradient descent optimizer with momentum factor $= 0.9$ and initial learning rate of $0.1$ is used. The learning rate is decayed by a factor of $10\times$ at 40, and 80 epochs. During the first 2 epochs the learning rate follows a linear warm-up schedule. $L2$ regularization with weight $4 \times 10^{-5}$ is used. Efficientnet \cite{tan2019efficientnet} preprocessing is used during training. Models are trained on 32 NVIDIA v100 GPUs with a per GPU batch size of 64. For all heteroscedastic models the optimal $\softmax$ temperature of $0.9$ is chosen based on the Imagenet ILSVRC12 optimal hyperparamters. A rank 15 low-rank approximation to the covariance matrix is used for our method. 1,000 MC samples are used at train and eval time for the heteroscedastic methods.

\paragraph{Imagenet-$21k$.} We train a ResNet-152v2 \cite{he2016deep} for 90 epochs. Stochastic gradient descent with momentum factor $= 0.9$ and initial learning rate of $0.1$ is used as the optimizer. The learning rate is decayed by a factor of $10\times$ at 30, 60 and 80 epochs. During the first 5000 steps the learning rate follows a linear warm-up schedule. $L2$ regularization with weight $1 \times 10^{-2}$ is used for the parameters mapping from the shared representation space to the low-rank covariance matrix, $L2$ regularization with weight $3 \times 10^{-3}$ is used for all other parameters. Standard data augmentation is used during training, an Inception crop followed by resizing to $224 \times 224$ and left-right horizontal flipping. All inputs are scaled to the $[-1, 1]$ range. Models are trained on $8 \times 8$ Google Cloud TPUv3s with a batch size of 1,024. Gradients are clipped to a maximum $L2$ of $1.0$. For all heteroscedastic models the optimal $\sigmoid$ temperature of $0.15$ is tuned on the validation set. A rank 50 low-rank approximation to the covariance matrix is used for our method. 1,000 MC samples are used at train and eval time for the heteroscedastic methods.

\paragraph{JFT.} We train a ResNet-50v2 \cite{he2016deep} for 30 epochs. Stochastic gradient descent with momentum factor $= 0.9$ and initial learning rate of $0.03$ is used as the optimizer. The learning rate is decayed by a factor of $10\times$ at 10, 20 and 25 epochs. During the first 5000 steps the learning rate follows a linear warm-up schedule. $L2$ regularization with weight $10^{-3}$ is used. Standard data augmentation is used during training, an Inception crop followed by resizing to $224 \times 224$ and left-right horizontal flipping. All inputs are scaled to the $[-1, 1]$ range. Models are trained on $16 \times 16$ Google Cloud TPUv3s with a batch size of 4,096. Gradients are clipped to a maximum $L2$ of $1.0$. For all heteroscedastic models the optimal $\sigmoid$ temperature of $0.15$ is chosen based on the Imagenet-$21k$ optimal hyperparamters. A rank 50 low-rank approximation to the covariance matrix is used for our method. 1,000 MC samples are used at train and eval time for the heteroscedastic methods.

\section{Analysis of the effect of the covariance matrix on the log-likelihood}
\label{app:correlations_effect}

We examine the effect of the covariance on the log-likelihood of our method, particularly in contrast to the homoscedastic log-likeihood. We make a 2\textsuperscript{nd} order Taylor series approximation to the log-likelihood of our method and show that the approximation decomposes into a term which corresponds to the homoscedastic log-likelihood and 2\textsuperscript{nd} order term which depends on the covariance matrix. We first examine the case when the covariance matrix is diagonal and then turn to the full covariance case.

For brevity denote the $\softmax(\mathbf{u})_k = \frac{\exp(u_k)}{\sum_j \exp(u_j)}$ as $s_k(\mathbf{u})$ which will sometimes abbreviate to $s_k$ when the $\softmax$ argument is clear. For simplicity we drop the dependence on the $\softmax$ temperature.

\paragraph{2\textsuperscript{nd} order Taylor series approximation.}

The second order Taylor series approximation to a single sample of the heteroscedastic output layer is given in Eq.~(\ref{eq:single_sample_approx}).

\begin{equation}
\begin{split}
    s_k(W^\intercal \mathbf{x} + V \boldsymbol{\epsilon}) &\approx s_k(W^\intercal \mathbf{x}) + \nabla s_k(W^\intercal \mathbf{x})^\intercal V \boldsymbol{\epsilon}\\ & + \frac{1}{2} \boldsymbol{\epsilon}^\intercal V^\intercal \nabla^2 s_k(W^\intercal \mathbf{x}) V \boldsymbol{\epsilon}
\end{split}
\label{eq:single_sample_approx}    
\end{equation}

where, $\boldsymbol{\epsilon} \sim \mathcal{N}(0_K, I_{K\times K})$.

Marginalizing over $\boldsymbol{\epsilon}$, the second term vanishes as $\mathbb{E} \brac{\boldsymbol{\epsilon}} = \boldsymbol{0}$. Hence the second order Taylor series approximation to the likelihood is:

\begin{equation}
\begin{split}
    \mathbb{E}_{\boldsymbol{\epsilon}} \brac{s_k(W^\intercal \mathbf{x} + V \boldsymbol{\epsilon})} &\approx s_k(W^\intercal \mathbf{x}) + \frac{1}{2} \textrm{tr}(\nabla^2 s_k(W^\intercal \mathbf{x}) V V^\intercal)
\end{split}
\label{eq:log_likelihood_approx}    
\end{equation}

\begin{lemma}
The Hessian of $s_k$, $\mathbf{H} s_k$, has the following structure:
\begin{equation*}
\left[\arraycolsep=1.8pt
\begin{array}{ccccc}
\ddots     &  & \cdots   & & \cdots\\
\cdots     & s_k (1 - s_k) (1 - 2 s_k)  & \cdots  & - s_j s_k (1 - 2 s_k) & \cdots \\
     & \ddots &  & & \\
     & & \ddots &  & \\
 2 s_k s_i s_j     & - s_j s_k (1 - 2 s_k)   & \cdots & - s_k s_j (1 - 2 s_j) & \cdots \\
\cdots     &       & \cdots  &   & \ddots
\end{array}
\right]
\end{equation*}
\end{lemma}

\paragraph{Proof.}
Note that the first derivative of the $s_k$ falls into two cases:
\begin{equation}
\frac{\partial s_k(\mathbf{u})}{\partial u_k} = \begin{cases}
- s_k s_j & k \neq j\\
 s_k (1 - s_k)  &  k = j.
\end{cases}
\end{equation}

\noindent
The second derivatives breakdown into four cases:

\textbf{Case 1}: $k \neq j, \frac{\partial^2 s_k(\mathbf{u})}{\partial u_j u_k} = - s_j s_k (1 - 2 s_k)$

\textbf{Case 2}: $\frac{\partial^2 s_k(\mathbf{u})}{\partial^2 u_k} = s_k (1 - s_k) (1 - 2 s_k)$

\textbf{Case 3}: $i, j \neq k, i \neq j, \frac{\partial^2 s_k(\mathbf{u})}{\partial u_i u_j} = 2 s_k s_i s_j$

\textbf{Case 4}: $j \neq k, \frac{\partial^2 s_k(\mathbf{u})}{\partial^2 u_j} = - s_k s_j (1 - 2 s_j)$

\subsection{Diagonal covariance}

Henceforth references to $s_k$ correspond to $s_k(W^\intercal \mathbf{x})$.

First, assume that the covariance matrix $V V^\intercal$ is diagonal with entries $\sigma_1^2, ..., \sigma_j^2, ..., \sigma_K^2$.

Substituting into Eq.~(\ref{eq:diag_likelihood_approx}) we get a special case of the approximate likelihood for the diagonal covariance case:

\begin{equation}
\begin{split}
    &\mathbb{E}_{\boldsymbol{\epsilon}} \brac{s_k(W^\intercal \mathbf{x} + V \boldsymbol{\epsilon})} \approx \\ &s_k(1 - \frac{1}{2} \sum_{j \neq k}^K s_j (1 - 2 s_j) \sigma_j^2 + \frac{1}{2} (1 - s_k) (1 - 2 s_k) \sigma_k^2)
\end{split}
\label{eq:diag_likelihood_approx}    
\end{equation}

With $\log(1 + t) \approx t$ the log-likelihood can be approximated as:

\begin{equation}
\begin{split}
    &\log \mathbb{E}_{\boldsymbol{\epsilon}} \brac{s_k(W^\intercal \mathbf{x} + V \boldsymbol{\epsilon})} \approx \\ & \log(s_k) -  \frac{1}{2} \sum_{j \neq k}^K s_j (1 - 2 s_j) \sigma_j^2 + \frac{1}{2} (1 - s_k) (1 - 2 s_k) \sigma_k^2
\end{split}
\label{eq:diag_log_likelihood_approx}    
\end{equation}

The $\log s_k = \log s_k(W^\intercal \mathbf{x})$ term in Eq.~(\ref{eq:diag_log_likelihood_approx}) is precisely the standard log-likelihood term of a homoscedastic model so the remaining $\frac{1}{2} \sum_{j \neq k}^K -s_j (1 - 2 s_j) \sigma_j^2 + \frac{1}{2} (1 - s_k) (1 - 2 s_k) \sigma_k^2$ term accounts for the approximate effect of the diagonal covariance on the heteroscedastic log-likelihood relative to the homoscedastic model.

Note that:

\begin{equation}
\begin{split}
    \frac{\partial}{\partial \sigma_j^2} \frac{1}{2} \sum_{j \neq k}^K -s_j (1 - 2 s_j) \sigma_j^2 + \frac{1}{2} (1 - s_k) (1 - 2 s_k) \sigma_k^2 \\ \propto -s_j (1 - 2 s_j)
\end{split}
\end{equation}

\begin{equation}
\begin{split}
    \frac{\partial}{\partial \sigma_k^2} \frac{1}{2} \sum_{j \neq k}^K -s_j (1 - 2 s_j) \sigma_j^2 + \frac{1}{2} (1 - s_k) (1 - 2 s_k) \sigma_k^2 \\ \propto (1 - s_k) (1 - 2 s_k)
\end{split}
\end{equation}

\begin{figure}
    \centering
    \includegraphics[width=0.45\textwidth]{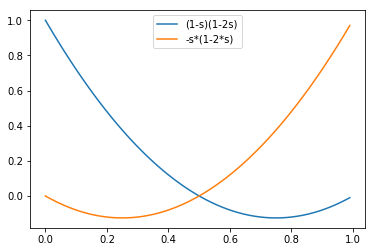}
    \caption{Behaviour of the coefficients that multiply the uncertainty factors in the second and third terms of the approximate likelihood, Eq. (\ref{eq:diag_log_likelihood_approx}).}
    \label{fig:likelihood_coefs}
\end{figure}

Fig.\ \ref{fig:likelihood_coefs} shows these derivatives. We see that when we observe a label $y = k$, then to maximize the log-likelihood, if $s_j > 0.5$ i.e., we are incorrectly classifying the example then the gradient forces $\sigma_j^2$ to increase and vice versa, if $s_j < 0.5$ the gradient encourages $\sigma_k^2$ to decrease. In this way a noisy label $y = k$ which may be confused with class $j$ can be explained away by a high $\sigma_j^2$ term.

Likewise if $s_k > 0.5$ i.e., we are correctly classifying the example then the gradient forces $\sigma_k^2$ to reduce and vice versa, when $s_k < 0.5$ the gradient encourages $\sigma_k^2$ to increase. Again the $\sigma_k^2$ allows the model to explain away a noisy label $y = k$ if the model assigns low probability to that label.

Interpreting the terms of the Taylor series relies on the 2nd order approximation being a reasonable approximation to the full method. We evaluate the efficacy of the approximation in the diagonal covariance case by training a ResNet-152 for 270 epochs using the 2nd order Taylor series approximation Eq.~\ref{eq:diag_likelihood_approx}. All hyperparameters are equivalent to those in the main paper. See Table~\ref{table:taylor_series_ilsvrc12} for results.

The deterministic approximate method recovers 1.5\% of the 2\% gains in top-1 accuracy seen by the full stochastic method. This is evidence that the Taylor series approximation is a reasonable approximation to the full method.

\begin{table}[tbh]
\centering
\begin{tabular}{lc}
\toprule
Method & Top-1 Acc \\
\midrule
Homoscedastic & 76.7 \\
Het.\ Diag & 78.7 \\ 
Het.\ Diag Taylor series approx & 78.2 \\ 
\bottomrule
\end{tabular}
\caption{Efficacy of 2nd order Taylor series approximation in diagonal covariance case for ResNet-152 trained on ILSVRC12 for 270 epochs.} 
\label{table:taylor_series_ilsvrc12}
\end{table}

\subsection{Full covariance}

If we allow a full covariance matrix $\Sigma = V V^\intercal$ then the approximate log-likelihood again breaks down into two terms, the first corresponding to the homoscedastic log-likelihood and the second corresponding to the effect of the heteroscedastic covariance matrix:

\begin{equation}
\begin{split}
    &\log \mathbb{E}_{\boldsymbol{\epsilon}} \brac{s_k(W^\intercal \mathbf{x} + V \boldsymbol{\epsilon})} \approx \log(s_k) \\ &  \frac{1}{2} \sum_{i \neq j, i,j \neq k}^K 2 s_i s_j \Sigma_{ij} + \frac{1}{2} \sum_{j \neq k}^K - s_j (1 - 2 s_k) \Sigma_{jk} \\
    & + \frac{1}{2} \sum_{j \neq k}^K - s_j (1 - 2 s_j) \Sigma_{jj} + \frac{1}{2} (1 - s_k) (1 - 2 s_k) \Sigma_{kk}
\end{split}
\label{eq:full_log_likelihood_approx}    
\end{equation}

The diagonal covariance terms $\Sigma_{kk}$ and $\Sigma_{jj}$ have the same interpretation to the additional terms in the diagonal covariance log-likelihood Eq.~(\ref{eq:diag_log_likelihood_approx}). So we will focus on the off-diagonal terms $\frac{1}{2} \sum_{i \neq j, i,j \neq k}^K 2 s_i s_j \Sigma_{ij} + \frac{1}{2} \sum_{j \neq k}^K - s_j (1 - 2 s_k) \Sigma_{jk}$.

$\frac{\partial}{\partial \Sigma_{ij}}\frac{1}{2} \sum_{i \neq j, i,j \neq k}^K 2 s_i s_j \Sigma_{ij} \propto s_i s_j$, so covariance matrix entries are encouraged to be large and positive when the product $s_i s_j$ is large i.e., when both classes $i$ and $j$ are assigned high probability by the homoscedastic term despite $y = k$. Classes $i$ and $j$ are encouraged to have a co-occurrence noise pattern.

\begin{figure}[t]
    \centering
    \includegraphics[width=0.45\textwidth]{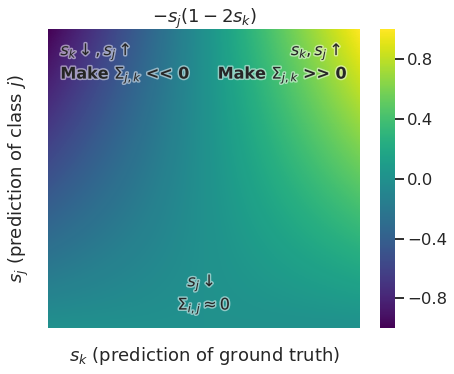}
    \caption{Heatmap of the derivatives of $\sum_{j \neq k}^K - s_j (1 - 2 s_k) \Sigma_{jk}$ w.r.t.\ $\Sigma_{jk}$. This is the effect of a small change in $\Sigma_{jk}$ when $j \neq k$ on the approximate log-likelihood, up to a constant multiplier, in the full covariance case. If the model is not predicting a high probability for (erroneous) class $j$, then there is no change to the noise term between $j$ and the correct class $k$. Otherwise, the model learns anticorrelations or correlations between them.}
    \label{fig:2d_coeffs_full_covariance}
\end{figure}

Likewise, we note the derivative of the $\frac{1}{2} \sum_{j \neq k}^K - s_j (1 - 2 s_k) \Sigma_{jk}$ term w.r.t.\ $\Sigma_{jk}$, $\frac{\partial}{\partial \Sigma_{jk}} \frac{1}{2} \sum_{j \neq k}^K - s_j (1 - 2 s_k) \Sigma_{jk} \propto - s_j (1 - 2 s_k)$. We see that to maximize the log-likelihood $\Sigma_{jk}$ is encouraged to be large and positive when both $s_j$ and $s_k$ are large (top right corner of Figure~\ref{fig:2d_coeffs_full_covariance}) and highly negative when either $s_j$ or $s_k$ is large and the other is small. So when we observe a label $y = k$ and classes $j$ and $k$ are assigned high probability then these classes are encouraged to have positively correlated noise i.e.\ to have a co-occurrence noise pattern. And when only one of the classes $j$ or $k$ has a high probability then the two classes are encouraged to have a substitution pattern in the noise. 

\section{Sensitivity to number of MC samples}
\label{app:mc_samples_sensitivity}

We test the sensitivity of our method to the number of MC samples. In Table~\ref{table:mc_sensitivty} we vary the number of training and test MC samples on Imagenet ILSVRC12, using the same experimental setup as ResNet-152 270 epoch results in the main paper. As expected increasing the number of MC samples improves performance monotonically; however there are diminishing returns beyond 100 samples.

\begin{table}[tbh]
\centering
\begin{tabular}{l|ccccc}
\toprule
\# MC Samples & 1 & 10 & 100 & 1000 & 10000 \\
\midrule
Top-1 ACC & 78.5 & 78.6 & 79.1 & 79.2 & 79.3 \\
\bottomrule
\end{tabular}
\caption{Sensitivity of ResNet-152 trained for 270 epochs on ILSVRC12 to the number of MC samples.}
\label{table:mc_sensitivty}
\end{table}

\section{Equalizing the parameter count}
\label{app:equal_params}

Our method adds additional parameters to the network. We investigate in this section whether the quality gains of our method are due to these additional parameters. The homoscedastic ResNet-152 model for ILSVRC12 has 60,185,128 parameters as compared to the full heteroscedastic method with 15 factors: 92,969,128. By equalizing the number of parameters for the homoscedastic model, we demonstrate that the gains from the heteroscedastic method are not primarily due to the additional parameters.

In particular we add an additional Dense + ReLU layer with 11,500 output units before the logits layer in a ResNet-152. This brings the total parameter count of the homoscedastic model to 93,200,628. However these additional parameters only lead to a +0.4\% increase in Top-1 accuracy, as opposed to +2.6\% with the full heteroscedastic method, Table~\ref{table:equal_params_ilsvrc12}. Therefore the gains we observe from using our method on Imagenet ILSVRC are due primarily to the noise modelling and not the increase in parameter count.

\begin{table}[tbh]
\centering
\begin{tabular}{lc}
\toprule
Method & Top-1 Acc \\
\midrule
Homoscedastic & 76.7 \\
Homoscedastic equal params & 77.1 \\
Het.\ Full (ours) & 79.3 \\ 
\bottomrule
\end{tabular}
\caption{Equalizing homoscedastic parameter count for ResNet-152 trained on ILSVRC12 for 270 epochs.} 
\label{table:equal_params_ilsvrc12}
\end{table}

\section{Training on ILSVRC12 with a sigmoid link function}
\label{app:sigmoid_ilsvrc12}

Prior work argues that ILSVRC12 is in reality a multilabelled dataset i.e.\ contains multiple objects per images, but that we force it to be a multiclass classification by assigning one of these objects as the ``primary'' object \cite{beyer2020we}. The authors then show that training a homoscedastic model on ILSVRC12 with the sigmoid output activation i.e.\ as if the ILSVRC12 labels were multilabelled leads to performance improvements.

However from a probabilsitic point of view this is an unsatisfying  solution. Using a sigmoid link function may yield improved accuracy but the network does not yield a valid probability distribution over the ILSVRC12 labels.

We have seen that the off-diagonal covariance matrix entries can model substitution patterns between co-occurring objects in a single image. We argue that the gains observed by \citet{beyer2020we} can be understood as arising from the misspecification of the homoscedastic model, due to the i.i.d.\ additive noise assumption.

In Table~\ref{table:sigmoid_ilsvrc12} we reproduce the gains from using the sigmoid link function for the homoscedastic model seeing the top-1 accuracy increase from 76.7\% to 78.2\%. However if we now train heteroscedastic model with the sigmoid link function, we see no significant gain compared to training the heteroscedastic model with a softmax link function. Thus the heteroscedastic noise model explains away the sigmoid effect. Removing the i.i.d.\ additive noise assumption has yielded a highly performant model which outputs a valid and well calibrated probability distribution over the observed labels.

\begin{table}[tbh]
\centering
\begin{tabular}{lc}
\toprule
Method & Top-1 Acc \\
\midrule
Homoscedastic (softmax) & 76.7 ($\pm 0.13$) \\
Homoscedastic (sigmoid) & 78.2 ($\pm 0.16$) \\
Het.\ Full (ours - softmax) & 79.3  ($\pm 0.10$) \\ 
Het.\ Full (sigmoid) & 79.35 ($\pm 0.05$) \\ 
\bottomrule
\end{tabular}
\caption{Training a ResNet-152 on ILSVRC12 with softmax vs.\ sigmoid link function.} 
\label{table:sigmoid_ilsvrc12}
\end{table}

\end{document}